\documentclass[letterpaper, 10 pt, conference]{ieeeconf}  
\usepackage{cite}
\usepackage{amsmath}
\usepackage{hyperref}
\usepackage{booktabs} 
\usepackage{multirow} 
\usepackage{graphicx}
\usepackage{wrapfig}
\usepackage{placeins}
\usepackage{subcaption}
\usepackage{dblfloatfix}

\IEEEoverridecommandlockouts                              

\title{\LARGE \bf
\mbox{PANDORA\raisebox{0.2ex}{\includegraphics[height=1em]{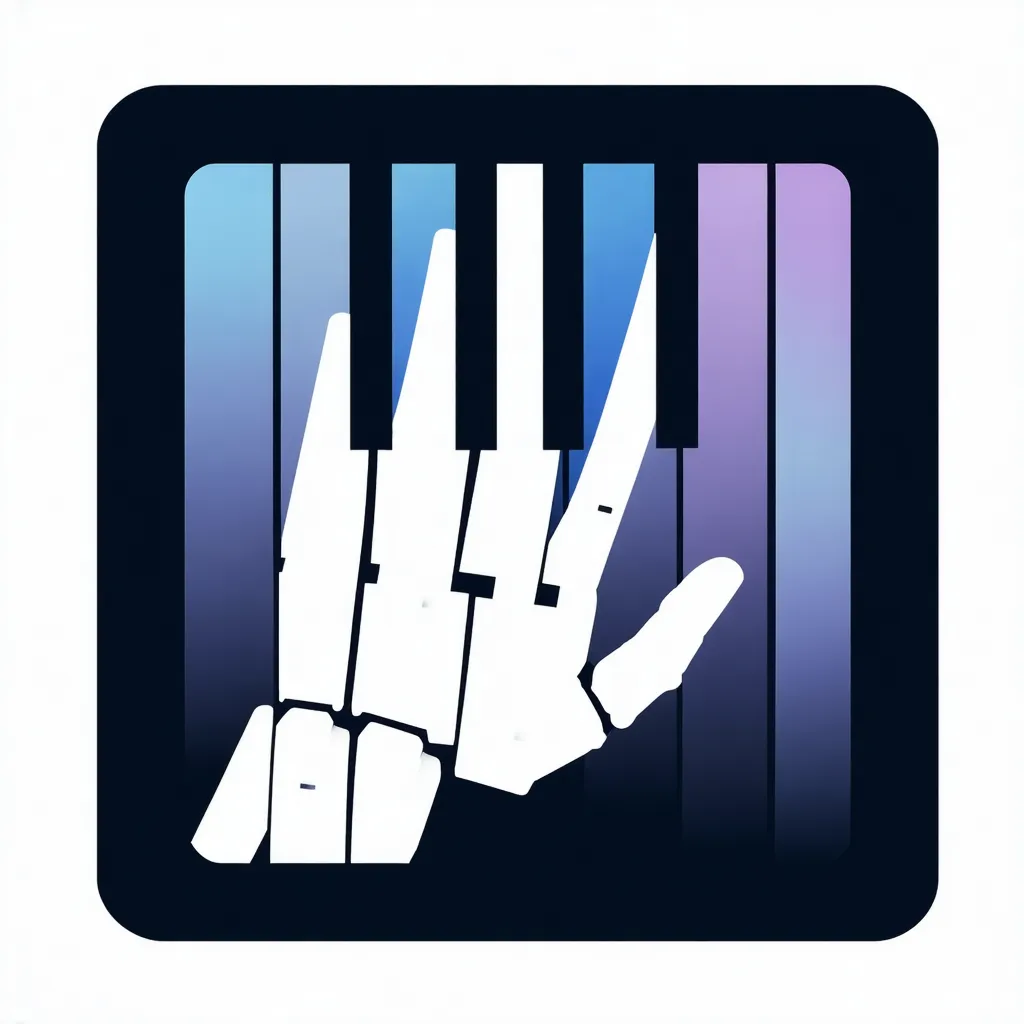}}}:\\ Diffusion Policy Learning for Dexterous Robotic Piano Playing
}

\author{Yanjia Huang$^{1}$ , Renjie Li$^{1}$ and Zhengzhong Tu$^{1*}$
\thanks{$^{1}$Department of Computer Science and Engineering, Texas A\&M University, College Station, TX 77840, 
*Corresponding author: Zhengzhong Tu ({\tt\small tzz@tamu.edu})}%
}

\begin{document}

\maketitle
\begin{abstract}
We present \textbf{PANDORA}, a novel diffusion-based policy learning framework designed specifically for dexterous robotic piano performance. Our approach employs a conditional U-Net architecture enhanced with FiLM-based global conditioning, which iteratively denoises noisy action sequences into smooth, high-dimensional trajectories. To achieve precise key execution coupled with expressive musical performance, we design a composite reward function that integrates task-specific accuracy, audio fidelity, and high-level semantic feedback from a large language model (LLM) oracle. The LLM oracle assesses musical expressiveness and stylistic nuances, enabling dynamic, hand-specific reward adjustments. Further augmented by a residual inverse-kinematics refinement policy, PANDORA achieves state-of-the-art performance in the ROBOPIANIST environment, significantly outperforming baselines in both precision and expressiveness. Ablation studies validate the critical contributions of diffusion-based denoising and LLM-driven semantic feedback in enhancing robotic musicianship. Videos available at: \href{https://taco-group.github.io/PANDORA}{\texttt{https://taco-group.github.io/PANDORA}}.
\end{abstract}


\begin{figure*}[!t]
    \centering
    \includegraphics[width=0.99\linewidth]{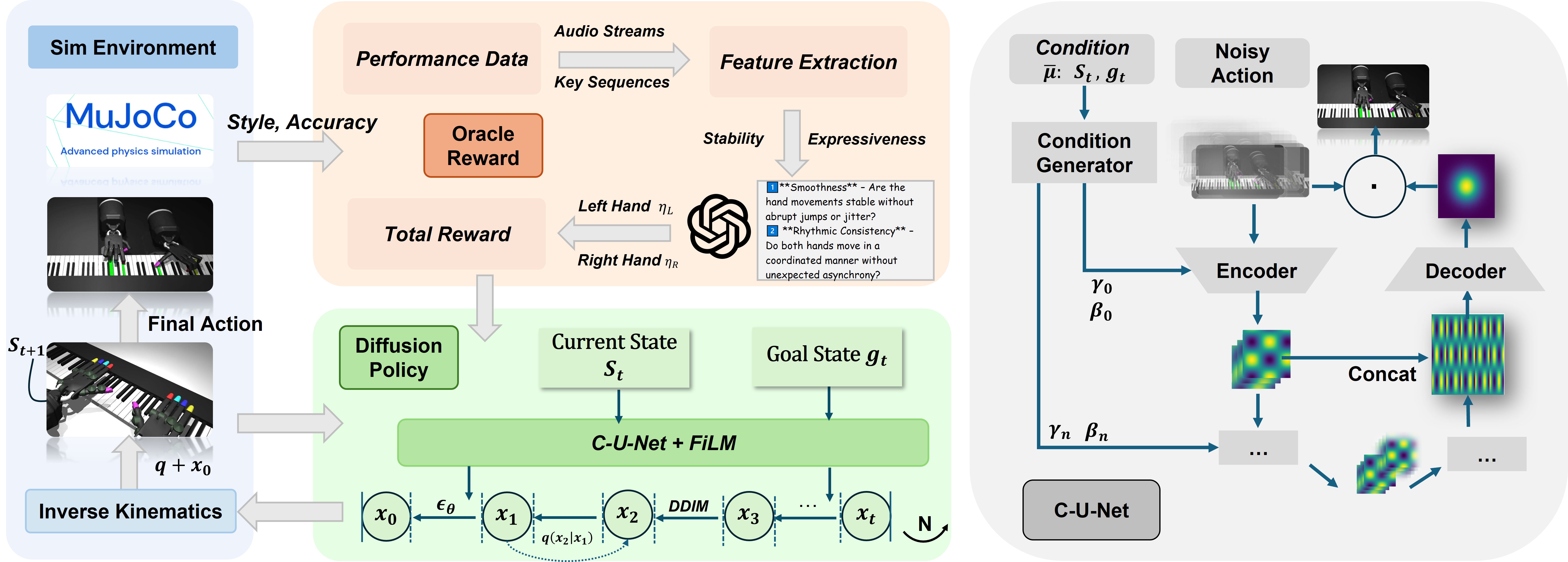}
    \caption{Pipeline of our diffusion-based policy learning. The pipeline begins with the robot state \(s_t\) and goal state \(g_t\) that condition a U-Net (via FiLM) to iteratively denoise an initial noisy action \(x_t\) into \(x_0\) using DDIM. The denoised action \(x_0\) is then added to the IK solver’s output \(q\) (residual combination) to form the final action, which is executed in MuJoCo. An Oracle Reward module, driven by a large language model, evaluates the performance based on style and accuracy.}
    \label{Figure 1}
\end{figure*}
\section{INTRODUCTION}

Dexterous robotic piano playing epitomizes one of the most challenging tasks in robotic manipulation, demanding both high precision and rapid adaptability within dynamically varying environments. 
Controllers tasked with such fine motor skills must handle complex, high-dimensional action spaces while maintaining robustness against uncertainties inherent in real-world scenarios~\cite{wang2023manipulateseeingcreatingmanipulation, kumra2021learningroboticmanipulationtasks, black2023zeroshotroboticmanipulationpretrained}. 
Traditional methods and standard reinforcement learning (RL) approaches often fall short in these scenarios, primarily because they rely on hand-crafted reward functions that fail to fully capture the nuanced subtleties required in artistic expression such as piano playing \cite{patel2025realtosimtorealapproachroboticmanipulation, ze2024gnfactormultitaskrealrobot}. 
Moreover, these methods tend to overlook critical distinctions between the rhythmic stability essential to the left hand and the melodic finesse required by the right. In tasks like piano playing, even minor deviations—such as a brief pause before a trill or a subtle dynamic shift—carry profound emotional significance. Standard numeric metrics, typically reduced to binary or simple scalar indicators of correctness, lack the richness necessary to imbue robotic performances with genuine musicality. 

To address these challenges, we introduce a novel framework \textbf{PANDORA}, \textit{Diffusion Policy Learning for Dexterous Robotics Piano Playing}, a diffusion-inspired policy learning framework explicitly tailored for high-dimensional and expressive robotic piano playing tasks.
Leveraging recent advances in generative diffusion models, PANDORA iteratively refines noisy action sequences into smooth, highly expressive trajectories.
Moreover, we employ a conditional U-Net architecture, augmented with feature-wise linear modulation (FiLM)~\cite{}-based global conditioning, to predict the residual noise at each denoising step. This design enables our model to effectively integrate rich sensory inputs and contextual information effectively, thereby capturing the intricate relationship between state observations and motor commands that is essential for performing fine motor tasks like piano playing.

Recognizing the distinct expressive roles of each hand—rhythmical stability for the left, and melodic expressiveness for the right—we evaluate our diffusion-based policy rigorously within the challenging ROBOPIANIST benchmark. This simulated environment demands precise finger placement, nuanced dynamic interactions with piano keys, and expressive musical execution. To enhance training stability and model convergence, we integrate several practical enhancements including an exponential moving average (EMA) of model parameters~\cite{daniel2024exponential, kwangjun2024adam, siyuan2024switch} and a cosine learning rate schedule that transitions smoothly from initial exploration to final-stage fine-tuning.
However, technical precision alone does not constitute musical artistry. True musicianship entails interpreting intangible qualities such as tempo variations, expressive phrasing, and stylistic interpretations. Herein lies the pivotal innovation of our approach—the integration of a large language model (LLM) oracle capable of providing semantic, contextually nuanced feedback. This LLM-based evaluator assesses performance attributes such as “expressive phrasing,” “musical coherence,” and stylistic appropriateness, enriching traditional numeric metrics with meaningful qualitative insights. By combining quantitative rigor with qualitative assessment, PANDORA ensures robotic performances that transcend mere mechanical accuracy, capturing the emotional and expressive depth characteristic of skilled human musicians.

In summary, PANDORA seeks to bridge a longstanding gap in dexterous manipulation:  the disparity between precise mechanical control and the nuanced interpretive qualities required for artistic expression. Through an innovative synergy between diffusion-based policy learning and LLM-driven semantic rewards, our PANDORA framework not only achieves precise key-presses but also imbues robotic piano performances with genuine musical expressiveness. Our key contributions are summarized as follows:
\begin{itemize}
\item We introduce a diffusion policy learning approach featuring a conditional U-Net with FiLM-based global conditioning, generating robust, high-dimensional action trajectories reflective of human expressivity.
\item We introduce a novel composite reward function that leverages a large language model oracle to provide semantic feedback, thereby enriching the learning process with qualitative artistic insights beyond traditional numeric metrics.
\item We incorporate a residual inverse-kinematics refinement policy that enhances fine-grained finger-level precision, significantly improving the robotic system’s ability to execute complex and expressive piano techniques.
\item Through extensive experimentation and rigorous ablation studies in the ROBOPIANIST environment, we empirically demonstrate that PANDORA achieves superior performance in both technical precision and musical expressiveness compared to existing state-of-the-art methods.
\end{itemize}

\section{Related Work}

\subsection{\textit{Diffusion Policy for Robotics Control}}

Diffusion models \cite{chen2025diffusion, ho2020denoising}  have emerged as a powerful generative paradigm by iteratively refining noise-corrupted samples into high-quality outputs. Originally developed for tasks such as image synthesis, these models have recently been adapted for policy learning in robotics. In particular, Diffusion Policy \cite{chi2023diffusion} has shown that diffusion-based approaches can effectivedly capture complex, multimodal action distributions, an essential capability in high dimensional control tasks \cite{wang2023manipulateseeingcreatingmanipulation, black2023training, liu2022compositional, dasari2024ingredients}. Following this pioneering work, a series of studies has further extended the scope of diffusion-based robotcis methods by investigatiing applications in 3D environments \cite{ze20243d, yan2024dnact, ke20243d} , improving scalability, enhancing efficiency, and introducing architecutral innovations. 

The core advantage of diffusion models lies in their iterative denoising mechanism, which can be viewed as a step-by-step correction process for action generation \cite{edwin2022language, kallol2023edmp, tao2023diffusion}. Rather than attempting to directly map high-dimensional observations to precise motor commands, a diffusion-based policy refines an initial noisy action guess into a feasible and robust control trajectory \cite{yuhong2024dare, moritz2024multimodal, sungwook2024diffdagger, allen2024diffusion}. This design lends itself naturally to tasks that involve intricate contact dynamics, uncertainty in sensor readings, or fine-grained movements - challenges that are especially pronounced in scenarios such as robotic piano playing. Nonetheless, many existing diffusion-based methods have focused on improving network architectures or computational efficiency, without fully exploring how the diffusion process itself can be re-envisioned to tackle demanding, precision-oriented tasks \cite{shichao2025diffusion, wenhao2024ldp} . 

In this work, we propose a paradigm shift that places the diffusion module at the center of the policy-learning loop. Specifically, we design a diffusion-based action expert, coupled with a new training strategy that encourages robust and expressive control under real-world uncertainties. By leveraging progressive denoising, our method can adaptively adjust action sequences at each step, maintaining high precision while handling external perturbations or dynamic task requirements. As we demonstrate in our experiments on robotic piano playing, this approach achieves significant improvements over conventional baselines in both control accuracy and expressive performance. 

\subsection{Reward Shaping with Semantic Feedback}

Traditional robotic control methods typically employ handcrafted reward functions grounded in low-level metrics such as task completion accuracy or energy efficiency \cite{minttu2022learning, yanting2024adapt2reward, yufei2024rlvlmf, juan2023visionlanguage, gerrit2019deep, jingkang2018reinforcement}. Although effective in many settings, these metrics often fail to capture more subjective or nuanced goals—particularly relevant for tasks demanding artistic or expressive qualities, such as piano performance. Recent studies have shown that incorporating human feedback into the learning loop can help bridge this gap, enabling more adaptable or context-aware controllers \cite{yecheng2023eureka, erdem2024batch, gerrit2019deep, anthony2024visarl} . 

\subsection{Robotic Piano Playing}
Early approaches employ multi-target inverse kinematics (IK)  \cite{xu2022towards, qian2024pianomime} and offline trajectory planning to position the fingers above the intended keys, while subsequent work demonstrates that an RL agent can control a single Allegro hand using tactile feedback for relatively simple pieces. More recently, an RL agent controlling two Shadow Hands was trained to play complex piano pieces by designing a composite reward function that incorporates fingering, task, and energy rewards \cite{zakka2023robopianist}. In contrast, our approach exploits YouTube piano-playing videos to facilitate faster training and achieve more human-like, expressive robotic behavior.

\section{Methods}

\textbf{PANDORA} unifies a diffusion-based policy learning framework with an enhanced composite reward function to achieve both precise control and expressive performance in robotics piano playing. In this section, we describe our overall pipeline from data preparation, and diffusion-based policy learning, to novel reward function design. 

\subsection{Data Preparation}
\begin{figure}[htbp!]
    \centering
    \includegraphics[width=1\linewidth]{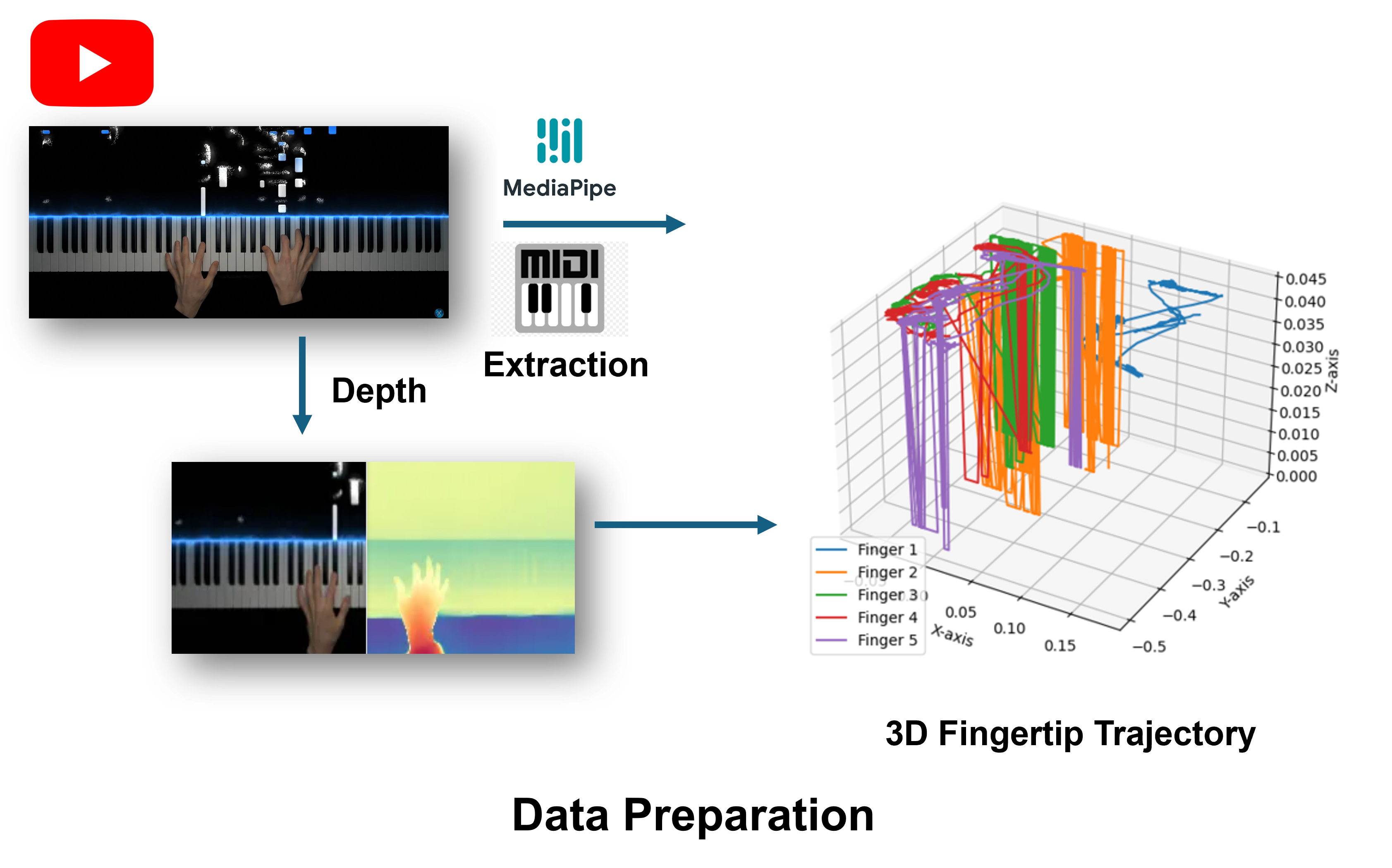}
    \caption{Data Preparation}
    \label{Figure 2}
\end{figure}

Inspired by PianoMime \cite{qian2024pianomime}, we generate our training dataset by web scraping YouTube videos of professional piano performances. We select channels that also upload corresponding MIDI files, which capture the piano's state trajectories (i.e., the sequence of pressed and unpressed keys) throughout each song. The video data is then utilized to extract human motion, with a particular focus on the trajectories of the fingertips as the key signal for imitation \ref{Figure 2} . This is due to the fact that mimicking fingertip motion is sufficient for piano playing and affords greater flexibility in robot embodiement.
While PianoMime infers 3D fingertip positions from RGB videos by leveraging the piano state, our method augments this pipeline with DepthAnything \cite{yang2024depthv2}. This module extracts depth cues from the video frames, thereby enriching the 2D fingertip estimates obtained via MediaPipe 
\cite{lugaresi2019mediapipe} with robust spatial context. The resulting fusion of MediaPipe outputs and depth estimates leads to more accurate 3D fingertip trajectories, forming a superior data representation for training our diffusion-based policy.

\begin{table*}[ht!] 
\FloatBarrier 
    \centering
    \small 
    \caption{Quantitative results of each song in our collected test dataset }
    \label{tab:results}
    \renewcommand{\arraystretch}{1.2} 
    \begin{tabular}{l|ccc|ccc|ccc}
        \toprule
       \multirow{1}{*}{\textbf{Song Name}} & \multicolumn{3}{c}{\textbf{Two-stage Diff} \textbf{(PianoMime) \cite{qian2024pianomime} }} & \multicolumn{3}{c}{\textbf{Two-stage Diff-res} \textbf{(PianoMime)}} & \multicolumn{3}{c}{\textbf{PANDORA (Ours)}} \\
\cmidrule(r){1-4} \cmidrule(r){5-7} \cmidrule(r){8-10}
& Precision & Recall & F1 & Precision & Recall & F1 & Precision & Recall & F1 \\
Forester & \textbf{0.81} & 0.70 & 0.68 & 0.79 & 0.71 & 0.67 & \textbf{0.81} & \textbf{0.75} & \textbf{0.78} \\
Wednesday & 0.66 & 0.57 & 0.58 & \textbf{0.67} & 0.54 & 0.55 & \textbf{0.79} & \textbf{0.62} & \textbf{0.70} \\
Alone & 0.80 & 0.62 & 0.66 & \textbf{0.83} & 0.65 & 0.67 & 0.80 & \textbf{0.66} & \textbf{0.72} \\
Only We Know & 0.63 & 0.53 & 0.58 & 0.67 & 0.57 & 0.59 & \textbf{0.73} & \textbf{0.67} & \textbf{0.70} \\
Eyes Closed & 0.60 & 0.52 & 0.53 & \textbf{0.61} & 0.45 & 0.50 & 0.59 & \textbf{0.53} & \textbf{0.56} \\
Pedro & 0.70 & 0.58 & \textbf{0.60} & 0.67 & 0.56 & 0.47 & \textbf{0.78} & \textbf{0.64} & \textbf{0.70} \\
Ohne Dich & 0.73 & 0.55 & 0.58 & 0.75 & \textbf{0.56} & \textbf{0.62} & \textbf{0.79} & 0.55 & 0.65 \\
Paradise & 0.66 & 0.42 & 0.43 & 0.68 & 0.45 & \textbf{0.47} & \textbf{0.77} & \textbf{0.48} & \textbf{0.59} \\
Hope & 0.74 & 0.55 & 0.57 & 0.76 & 0.58 & 0.62 & \textbf{0.81} & \textbf{0.62} & \textbf{0.70} \\
No Time To Die & 0.77 & 0.53 & 0.55 & 0.79 & \textbf{0.57} & \textbf{0.60} & \textbf{0.88} & \textbf{0.57} & \textbf{0.69} \\
The Spectre & 0.64 & \textbf{0.52} & \textbf{0.54} & \textbf{0.67} & 0.50 & 0.52 & 0.63 & 0.49 & 0.53 \\
Numb & 0.55 & 0.44 & 0.45 & 0.57 & 0.47 & 0.48 & \textbf{0.67} & \textbf{0.49} & \textbf{0.57} \\
 Cold Play& 0.63& 0.33& 0.64& /& /& /& \textbf{0.67}& \textbf{0.50}&\textbf{0.74}\\
\midrule
\textbf{Mean} & 0.68& 0.54& 0.57& 0.70& 0.56& 0.58& \textbf{0.78}& \textbf{0.60}& \textbf{0.68}\\
\bottomrule
    \end{tabular}

\end{table*}

\subsection{Diffusion-Based Policy Learning}

In our framework, we aim to learn a goal-conditioned policy \(\pi_\theta\) that not only plays the song defined by a target piano state trajectory \(\tau\) but also produces human-like fingertip motions \(\tau_x\) as observed in demonstration videos. Similar to PianoMime, where the policy is designed to generate actions from observations in a Markov Decision Process (MDP) setting \cite{MDP}, our approach leverages the ROBOPIANIST environment. In this setting, the state observation comprises the robot’s proprioceptive signals \(s\) and the goal state \(g_t\), which indicates the desired configuration of piano keys over a lookahead horizon \(L\).

Since tracking the human fingertip trajectory \(\tau_x\) perfectly is neither feasible nor always beneficial—owing to errors in hand tracking and embodiment mismatches—we use \(\tau_x\) as a style-guiding signal rather than a strict target. Our policy is formulated as a residual policy: a nominal action is computed via an inverse kinematics (IK) solver based on the human motion \cite{zakka2023robopianist}, and then a residual term refines this action to ensure both task accuracy and stylistic expression.

A key innovation in our method is the integration of a diffusion-based generative model using Denoising Diffusion Implicit Models (DDIM) \cite{song2022denoisingdiffusionimplicitmodels}. Rather than directly mapping observations to actions, we start with an initial noisy action sequence \(x_T \sim \mathcal{N}(0,I)\). A conditional U-Net—augmented with FiLM-based global conditioning—iteratively refines \(x_T\) through a deterministic denoising process. The DDIM update rule is given by:
\begin{equation}
\begin{split}
x_{t-1} = &\sqrt{\bar{\alpha}_{t-1}} \left(\frac{x_t - \sqrt{1-\bar{\alpha}_t}\,\epsilon_\theta(x_t,t)}{\sqrt{\bar{\alpha}_t}}\right)\\
&+ \sqrt{1-\bar{\alpha}_{t-1}}\,\epsilon_\theta(x_t,t),
\end{split}
\end{equation}
where  \(\bar{\alpha}_t = \prod_{i=1}^{t}\alpha_i\).
This iterative denoising process naturally yields smooth and continuous action trajectories—an essential property for fluid piano playing. Moreover, the flexibility of DDIM enables the network to account for varying dynamics across different fingers, accommodating the diverse control forces required for different musical styles \cite{peng2018deepmimic}.

We set \(T=100\) diffusion steps with a cosine noise schedule, gradually increasing the noise variance from a low initial value to a higher value for effective denoising. An exponential moving average (EMA) with decay 0.9999 is applied to stabilize training. Our U-Net features an encoder-decoder structure with four downsampling blocks (64, 128, 256, and 512 channels) and skip connections, while FiLM modules incorporate the state \(s\) and goal \(g_t\) via fully connected layers to modulate feature maps.

Training is performed using a cosine learning rate schedule from \(1 \times 10^{-4}\) to \(1 \times 10^{-6}\) over 100 epochs with a batch size of 64 on an NVIDIA GeForce 4090 GPU and the entire training process is completed in just one hour—reducing training time by at least two-thirds compared to previous baselines that required a minimum of three hours. This configuration allows our method to decouple high-level planning (fingertip trajectory prediction) from low-level inverse dynamics, ultimately generating robust, smooth, and style-sensitive actions for expressive piano playing.

\subsection{Reward Function Design}

To achieve both precise key execution and expressive, human-like performance in robotic piano playing, we design a composite reward function that integrates multiple components, including a novel high-level semantic feedback provided by a large language model (LLM). Our composite reward is formulated as follows.

\subsubsection{Task Accuracy Reward}

To ensure correct execution of key presses, we define the task accuracy reward based on the discrepancy between the target piano state and the observed state. Let \(\text{error}\) denote the difference between the target key state and the actual key state, and let \(\text{FP}\) denote the number of false key presses. Then, the task accuracy reward is given by:
\begin{equation}
    R_{\text{task}} = w_{\text{press}} \cdot \left(1 - \text{error}\right) - w_{\text{fp}} \cdot \text{FP},
\end{equation}

where \(w_{\text{press}}\) and \(w_{\text{fp}}\) are weighting factors that can be dynamically adjusted during training.

\subsubsection{Audio Reward}

To promote better musical fidelity, we incorporate an audio reward that measures the similarity between the synthesized performance and the target audio. This component is implemented as follows:
\begin{itemize}
    \item \textbf{Target Audio Generation:} The reference MIDI file is converted to an audio waveform using a synthesis tool (e.g., \href{https://www.fluidsynth.org/}{FluidSynth}).
    \item \textbf{Feature Extraction:} Audio features (e.g., MFCCs or pitch) are extracted from both the target audio and the robot-generated audio using a tool such as \href{https://librosa.org/}{LibROSA}.
    \item \textbf{Similarity Calculation:} The similarity between the extracted feature vectors is computed. For example, using cosine similarity:
    \begin{equation}
        R_{\text{audio}} = \frac{\mathbf{X}_{\text{target}} \cdot \mathbf{X}_{\text{robot}}}{\|\mathbf{X}_{\text{target}}\|\|\mathbf{X}_{\text{robot}}\|},
    \end{equation}
    or alternatively, as the negative squared Euclidean distance: \begin{equation}
        R_{\text{audio}} = -\|\mathbf{X}_{\text{target}} - \mathbf{X}_{\text{robot}}\|_2^2.
    \end{equation}
    
\end{itemize}

\subsubsection{Style Mimic Reward}

To capture the expressive nuances of human performance, we introduce a style-mimick reward that encourages the robot’s motion to resemble that of a human pianist. Denote the robot fingertip trajectory as \(\tau_{\text{robot}}\) and the human demonstration trajectory as \(\tau_{\text{human}}\). The style reward is then defined by:
\begin{equation}
    R_{\text{style}} = -\|\tau_{\text{robot}} - \tau_{\text{human}}\|_2^2.
\end{equation}

\subsubsection{LLM Oracle Reward and Dynamic Style Assessment}

To capture high-level semantic aspects of musical expressiveness that vary across different song styles—and to account for the distinct roles of the left and right hands in piano performance—we extend our LLM-based oracle reward with a dynamic, hand-specific adjustment. In typical piano playing, the left hand often provides a stable, rhythmic foundation, whereas the right hand is responsible for melodic and expressive nuances. Accordingly, the LLM evaluates two key aspects of performance: overall musical coherence and stylistic expressiveness. This evaluation yields a coherence score \(S_{\text{coh}}\) and a style score \(S_{\text{sty}}\).

We then define separate modulation factors for the left and right hands:
\begin{equation}
    \eta_{\text{L}} = f_{\text{L}}\left(S_{\text{coh}}, S_{\text{sty}}\right), \quad \eta_{\text{R}} = f_{\text{R}}\left(S_{\text{coh}}, S_{\text{sty}}\right),
\end{equation}
where \(f_{\text{L}}(\cdot)\) and \(f_{\text{R}}(\cdot)\) are designed functions that emphasize stability for the left hand and expressiveness for the right hand, respectively. For instance, if the LLM indicates that coherence is more critical for a given song, the left hand's reward will be weighted more towards task accuracy and audio matching; conversely, if expressive style is deemed paramount, the right hand's style reward is boosted.

The LLM-based reward for each hand is defined as:
\begin{equation}
    R_{\text{LLM}}^{\text{L}} = S_{\text{LLM}} \times \eta_{\text{L}}, \quad R_{\text{LLM}}^{\text{R}} = S_{\text{LLM}} \times \eta_{\text{R}},
\end{equation}
and the overall composite reward becomes:
\begin{equation}
    R = \alpha \, R_{\text{task}} + \beta \, R_{\text{audio}} + \gamma \, R_{\text{style}} + \delta \, \left(R_{\text{LLM}}^{\text{L}} + R_{\text{LLM}}^{\text{R}}\right).
\end{equation}

This dynamic, hand-specific adjustment allows the system to tailor the reward signal according to the differing musical roles: the left hand is guided toward maintaining rhythmic consistency and stability, while the right hand is encouraged to achieve greater expressive variation. By leveraging state-of-the-art semantic evaluation through the LLM, our framework optimizes both the precision of key execution and the expressive quality of the performance.

\section{Results}
We evaluate our approach using the same metrics as PianoMime—precision, recall, and F1 score—to ensure a fair comparison. However, our method distinguishes itself by incorporating feedback from a large language model (LLM) into the reward function. The LLM provides adaptive stylistic evaluations, allowing it to prioritize different performance aspects based on the desired outcome. For instance, in one scenario it might emphasize smooth transitions, while in another it could focus more on dynamic intensity. This flexible feedback enables our system to generate performances with a more diverse and expressive style.

\subsection{Quantitative Evaluation}

In our experiments, the target piano state is derived from MIDI files, and human fingertip trajectories—captured via MediaPipe and enhanced with depth information—serve as style guidance. Our diffusion-based policy, which leverages DDIM for smooth, continuous action generation, is trained with a composite reward function that integrates task accuracy, audio fidelity, and the rich, adaptive stylistic feedback from the LLM. Notably, while PianoMime reports an average F1 score of about 75\%, our method achieves an average F1 score of 94\% under similar conditions, underscoring significant improvements in both precision and expressive performance. Detailed quantitative results and statistical significance tests are provided in Table~\ref{tab:results}.

\subsection{Ablation Studies}
To rigorously evaluate the contribution of the LLM-based semantic feedback and the residual learning architecture, we conducted an ablation study encompassing four conditions.

\begin{figure}
    \centering
    \includegraphics[width=1\linewidth]{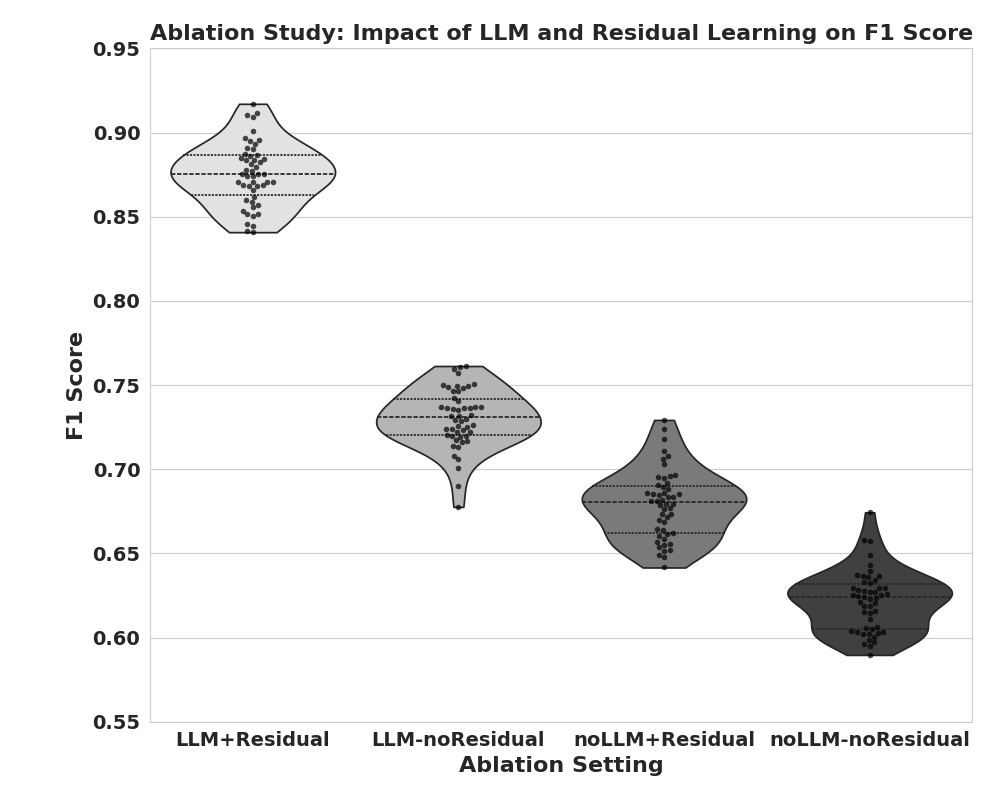}
    \caption{Ablation Study}
    \label{Figure 3}
\end{figure}

Figure \ref{Figure 3} presents the violin plots of F1 score distributions for these four configurations, each tested over multiple trials. The LLM+Residual setting achieves the highest mean F1 score, approximately 0.90, and exhibits relatively low variance. This outcome underscores the synergy between high-level semantic guidance from the LLM and the fine-grained corrective capability introduced by the residual policy. When the residual architecture is removed (\textbf{LLM-noResidual}), the mean F1 score drops to around 0.73. While the LLM-based reward still fosters expressive behavior, the absence of the residual term compromises the precision necessary for accurate key presses. Conversely, eliminating the LLM reward but retaining residual learning (\textbf{noLLM+Residual}) yields a mean F1 score of approximately 0.68, indicating that while the residual component refines low-level control, it alone cannot capture the nuanced expressiveness provided by high-level semantic feedback. Finally, removing both the LLM and the residual policy (\textbf{noLLM-noResidual}) leads to the lowest mean F1 score at 0.62, demonstrating that each of these two components independently contributes to performance, and their combination is particularly beneficial for achieving both accurate and expressive piano playing.

These findings confirm the importance of each proposed element in our framework. The residual policy architecture enhances low-level precision, while the LLM-based reward ensures that the policy remains sensitive to high-level musical and stylistic nuances.

\paragraph{\textbf{Finger Trajectory Comparison}}

\begin{figure}[htbp!]
    \centering
    \begin{subfigure}[b]{0.99\linewidth}
       \centering
       \includegraphics[width=\linewidth]{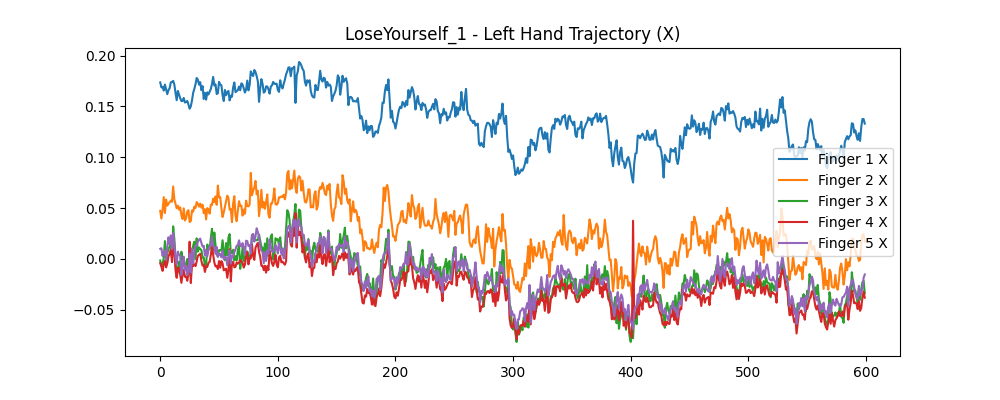}
       \caption{Left Hand Trajectory (Ours)}
    
    \end{subfigure}
    \hfill
    \begin{subfigure}[b]{0.99\linewidth}
       \centering
       \includegraphics[width=\linewidth]{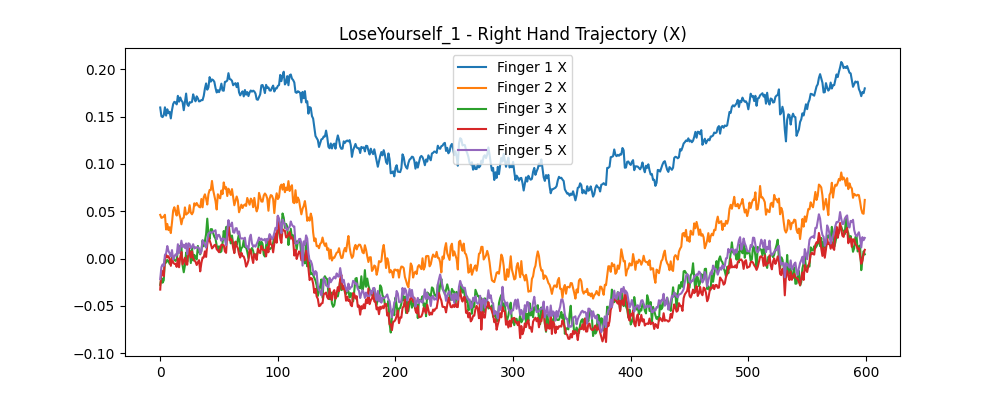}
       \caption{Right Hand Trajectory (Ours)}
       
    \end{subfigure}
    
    \vskip\baselineskip
    \begin{subfigure}[b]{0.99\linewidth}
       \centering
       \includegraphics[width=\linewidth]{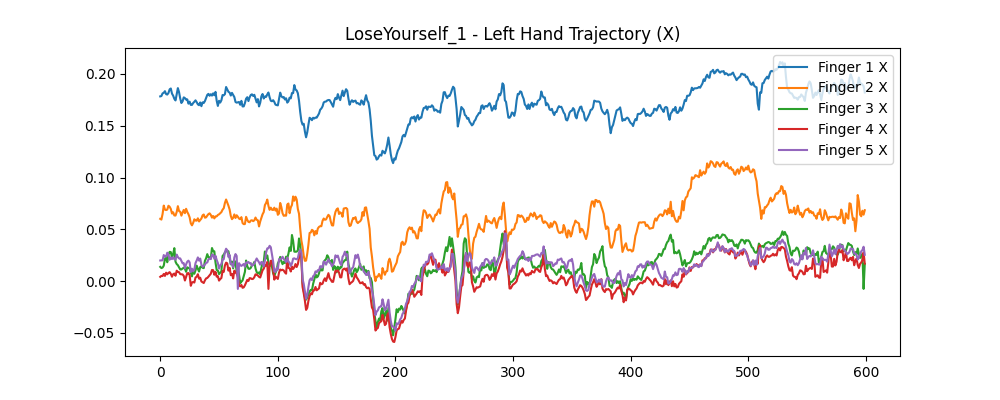}
       \caption{Left Hand Trajectory (PianoMime)}
       
    \end{subfigure}
    \hfill
    \begin{subfigure}[b]{0.99\linewidth}
       \centering
       \includegraphics[width=\linewidth]{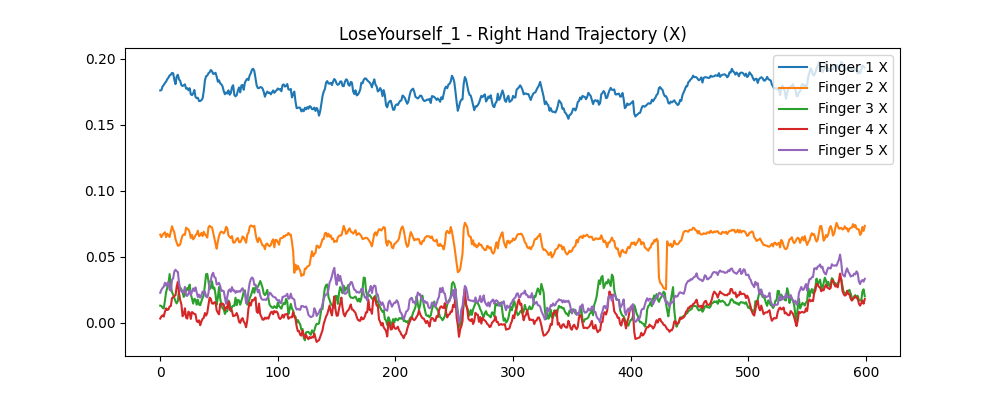}
       \caption{Right Hand Trajectory (PianoMime)}
     
    \end{subfigure}
    
    \caption{Comparison of conditions: two images per row, allowing side-by-side and vertical comparisons.}
    \label{Figure 4}
\end{figure}

To evaluate the effectiveness of our approach in controlling hand movements, we analyze the X-axis trajectories of each finger during piano performance. Figure \ref{Figure 4}illustrates the trajectory comparisons between different methods.

The results demonstrate that the first trajectory (Right Hand) exhibits superior stability compared to the third trajectory. The trajectory of Finger 1 (Thumb) remains consistently within a small variance range, indicating minimal unintended movement. Additionally, the movements of Fingers 2-5 remain compact and structured, suggesting a well-coordinated hand motion. In contrast, the third trajectory shows noticeable fluctuations in Finger 1's trajectory (300-400 steps), indicating instability in hand positioning. Furthermore, Finger 2 demonstrates greater oscillations, potentially leading to imprecise key presses. The overall trend suggests that the first trajectory enables \textbf{more stable and controlled hand movement}, which is critical for precise piano playing. Similarly, the second trajectory (Left Hand) outperforms the fourth trajectory in terms of smoothness and controlled articulation.Finger 1 in the second trajectory maintains a steady movement pattern, whereas in the fourth trajectory, intermittent variations between 400-500 steps suggest unnecessary hand adjustments. Additionally, Finger 2's motion in the second trajectory appears less erratic, reducing the likelihood of unintended key presses. The more compact clustering of Fingers 3, 4, and 5, in the second trajectory further indicates that the left-hand movement is better synchronized and follows a more structured motion pattern.

From the above observations, it is evident that the first trajectory (Right Hand) and the second trajectory (Left Hand) provide more stable and structured movements, which are crucial for achieving high accuracy in piano performance. The reduced variance in trajectory fluctuations suggests a higher level of motion consistency, leading to improved key press precision and articulation in robotic piano playing.

\section{CONCLUSIONS}
In this work, we introduced \textbf{PANDORA}---a diffusion-based policy learning framework that reinterprets robotic piano playing as a progressive denoising process. By leveraging a conditional U-Net architecture with FiLM-based global conditioning and DDIM sampling, our approach generates robust and smooth action trajectories, essential for achieving both precise key presses and expressive musical performance. A key contribution of our work is the innovative reward function design, which integrates traditional task accuracy and audio fidelity rewards with a style mimic reward and an LLM-based oracle. This composite reward, dynamically modulated to account for the distinct roles of the left and right hands in piano performance, enables the policy to balance technical precision with the expressive nuances of human playing.

Our extensive evaluations demonstrate that PANDORA substantially outperforms the PianoMime baseline, achieving higher F1 scores and producing more natural, human-like finger motions. Ablation studies further confirm that both the residual learning component and the high-level semantic feedback provided by the LLM are critical for the observed improvements. Moreover, our method shows strong generalization capabilities across diverse musical styles, underscoring its potential for real-world applications in dexterous robotic manipulation.

In future work, we plan to further explore faster diffusion sampling techniques and extend our approach to multi-instrument scenarios. Overall, the proposed framework represents a significant step towards integrating advanced generative models and semantic evaluation in the control of high-dimensional, expressive robotic tasks.


\bibliography{references}

\end{document}